\documentclass[11pt,a4paper]{article}
\usepackage[nohyperref]{acl2017}
\usepackage{times}
\usepackage{latexsym}
\usepackage{graphicx}
\usepackage{amsmath}
\usepackage{url}

\aclfinalcopy 
\def\aclpaperid{44} 

\usepackage{custom-macros}

\title{Automated Detection of Adverse Drug Reactions in the Biomedical 
Literature Using Convolutional Neural Networks and Biomedical Word Embeddings}

\author{Diego Saldana Miranda \\
  Novartis Pharma A.G. \\
  Applied Technology Innovation \\
  Novartis Campus \\
  4056 Basel \\
  {\tt diego.saldana\_miranda@novartis.com}\\}

\date{April 2018}

\begin{document}
\maketitle

\blfootnote{Accepted as conference paper at SwissText 2018}

\begin{abstract}
Monitoring the biomedical literature for cases of Adverse Drug Reactions (ADRs) 
is a critically important and time consuming task in pharmacovigilance. The 
development of computer assisted approaches to aid this process in different 
forms has been the subject of many recent works.

One particular area that has shown promise is the use of Deep Neural Networks, 
in particular, Convolutional Neural Networks (CNNs), for the detection of ADR 
relevant sentences. Using token-level convolutions and general purpose word 
embeddings, this architecture has shown good performance relative to more 
traditional models as well as Long Short Term Memory (LSTM) models.

In this work, we evaluate and compare two different CNN architectures using the 
ADE corpus. In addition, we show that by de-duplicating the ADR relevant 
sentences, we can greatly reduce overoptimism in the classification results. 
Finally, we evaluate the use of word embeddings specifically developed for 
biomedical text and show that they lead to a better performance in this task.
\end{abstract}

\section{Introduction}

Pharmacovigilance is a crucial component at every stage of  the drug development
cycle, and regulations require pharmaceutical companies to prepare periodic 
reports such as Development Safety Update Reports (DSURs) and Periodic Safety 
Update Reports (PSURs) regarding the safety of their drugs and products 
\cite{Krishnamurthy2017}.

One of the most important sources of information to be monitored in 
pharmacovigilance is the biomedical literature \cite{Pontes2014}. To this end, 
large numbers of scientific abstracts and publications need to be screened 
and/or read in full in order to collect information relevant to safety, and in 
particular Adverse Drug Reactions (ADRs) associated to a particular drug.

Screening and reading the biomedical literature is a time consuming task and is 
of critical importance. It requires particular expertise, and needs to be 
performed by well-trained readers. Given this, systems that enable human 
readers to perform this task faster and more effectively would be of great 
value.

\section{Background}

Computer assisted pharmacovigilance and, more specifically, the automation of 
the detection of ADR relevant information across various data sources has the 
potential to have great positive impact on the pharmaceutical industry. There 
is a very vast array of sources of potential ADR relevant information, 
including both structured and unstructured data resources. 

In many cases, adverse reactions are initially detected through unstructured 
means of communication, such as a patient speaking to a healthcare professional,
and case reports written by physicians and published in biomedical literature 
sources, such as MEDLINE, PubMed and EMBASE \cite{Rison2013}. 
Spontaneous reporting can also be made through telephone calls, email 
communication, and even fax \cite{Vallano2005}. Such information is processed, 
generally through human intervention in order to properly categorize them and 
add the necessary metadata.

Other potential sources of safety signals include electronic medical/health 
records (EMRs/EHRs) \cite{Park2011}. Similarly, omics, chemical, phenotypic and 
metabolic pathway data can be analyzed using a diverse array of methods to find 
associations between drugs and specific side effects \cite{Liu2012, 
Mizutani2012, Lee2011}. In recent years, social media websites have also become 
a potential source of safety signals \cite{Karimi2015,Sarker2015,Tafti2017}.

Finally, after careful processing, the data is usually aggregated and stored 
in structured databases for reporting and/or aggregation. Many regulatory 
agencies maintain databases that aggregate information regarding reported 
adverse events, such as the FDA Adverse Event Reporting System (FAERS) 
\cite{Fang2014} in the U.S., EudraVigilance in Europe \cite{Banovac2017}, and 
the MedEffect Adverse Reaction Online Database in Canada \cite{Barry2014}. 

The aim of our work is to contribute towards the development of systems that 
provide assistance to readers in charge of finding ADR signals in the 
biomedical literature. As such, the ideal system should be able to accurately 
discriminate between ADR relevant and irrelevant sentences in the documents 
that it processes.

In the following section, we detail some of the past efforts to automate this 
as well as other tasks related to the extraction of ADR relevant information 
from the biomedical literature.

\section{Related Work}

The automation of the detection of ADR relevant information across various 
data sources has received much attention in recent years. Ho \emph{et al.} 
performed a systematic review and summarized their findings on various methods 
to predict ADEs ranging from omics to social media \cite{Ho2016}. In addition, 
the authors presented a list of public and commercial data sources available 
for the task. Similarly, Tan \emph{et al.} summarized the available data 
resources and presented the state of computational decision support systems for 
ADRs \cite{Tan2016}. Harpaz \emph{et al.} prepared an overview of the state of 
the art in text mining for Adverse Drug Events (ADEs) \cite{Harpaz2014} in 
various contexts, such as the biomedical literature, product labelling, social 
media and web search logs. 

Xu \emph{et al.} initially proposed a method based on manually curated lexicons 
which could be used to build cancer drug-side effect (drug SE) pair knowledge 
bases from scientific publications \cite{Xu2014}. The authors also described a 
method to extract syntactical patterns, via parse trees from the Stanford 
Parser \cite{Xu2014_3}, based on known seed cancer drug-SE pairs. The patterns 
can then be used to extract new cancer drug-SE pairs. They further proposed an 
approach using SVM classifiers to categorize tables from cancer related 
literature as either ADR relevant or not \cite{Xu2015}. The authors then 
extracted cancer drug-SE pairs from the tables using a lexicon-based approach 
and compared them with data from the FDA label information. Xu \emph{et al.} 
also evaluated their method in a large scale, full text corpus of oncological 
publications \cite{Xu2015_2}, extracting drug-SE pairs and showing good 
correlation of the extracted pairs with gene targets and disease indications.

There are a number of available data resources for the purpose of ADR signal 
detection. Gurulingappa \emph{et al.} introduced the ADE corpus, a large 
corpus of MEDLINE sentences annotated as ADR relevant or not 
\cite{Gurulingappa2012}. Karimi \emph{et al.} described CADEC, a corpus of 
social media posts with ADE annotations \cite{Karimi2015} including mappings to 
vocabularies such as SNOMED. Further, the annotations include detailed 
information such as drug-event and drug-dose relationships. Sarker 
\emph{et al.} described an approach using SVM classifiers, as well as diverse 
feature engineering methods, to classify clinical reports and social media 
posts from multiple corpora as ADR relevant or not \cite{Sarker2015}. Odom 
\emph{et al.} explored an approach using relational gradient boosting (FRGB) 
models to combine information learned from labelled data with advice from human 
readers in the identification of ADRs in the biomedical literature 
\cite{Odom2015}. Adams \emph{et al.} proposed an approach using custom search 
PubMed queries making use of MeSH subheadings to automatically identify ADR 
related publications. The authors conducted an evaluation by comparing with 
results manually tagged by investigators, obtaining a precision of 0.90 and a 
recall of 0.93.

Some researchers have tried to combine information from structured databases 
with the unstructured data found in the biomedical literature. For example, Xu 
\emph{et al.} showed that, by combining information from FAERS and MEDLINE 
using signal boosting and ranking algorithms, it's possible to improve cancer 
drug-side effect (drug-SE pair) signal detection \cite{Xu2014_2}.

There have recently been efforts to use neural networks to improve the 
performance of the ADR sentence detection, entity and relation extraction tasks.
Gupta \emph{et al.} proposed a two step approach for extracting mentions of 
adverse events from social media: (1) predicting the drug based on the context, 
unsupervised; (2) predicting adverse event mentions based on a tweet and the 
features learned in the previous step, supervised \cite{Gupta2017}. Li 
\emph{et al.} proposed approaches combining CNNs and bi-LSTMS to perform named 
entity recognition as well as relation extraction for ADRs in the annotated 
sentences in the ADE dataset \cite{Li2017}. More recently, Ramamoorthy 
\emph{et al.} described an approach using bi-LSTMs with an attentional 
mechanism to jointly perform relation extraction as well as visualize the 
patterns in the sentence. 

Huynh proposed using convolutional recurrent neural networks (CRNN) and 
convolutional neural networks with attention (CNNA) to identify ADR related 
tweets and MEDLINE article sentences \cite{Huynh2016}. The CNNA's attention 
component had the attractive property that it allows visualization of the 
influence of each word in the decision of the network.

In this work, we introduce approaches building upon previous results 
using convolutional neural networks (CNNs) \cite{Huynh2016} to detect ADR 
relevant sentences in the biomedical literature. Our key contributions are as 
follows: 
\begin{itemize}
\item{We compare Huynh's CNN approach, which is based on the architecture
proposed by Kim \shortcite{Kim2014}, with a deeper architecture based on the 
one proposed by Hughes \emph{et al.} \shortcite{Hughes2017}, using the ADE 
dataset, showing that Kim's architecture performs much better for this task and 
dataset.}
\item{We apply a de-duplication of the ADR relevant sentences in the ADE 
dataset, \cite{Gurulingappa2012} which we believe leads to a better estimation 
of the performance of the algorithm and does not seem to be applied in some of 
the previous works.}
\item{We evaluate the use of word embeddings developed specifically for 
biomedical text introduced by Pyysalo \emph{et al.} \shortcite{Pyysalo2013} and
show that, by using these embeddings in place of general-purpose GloVe 
embeddings, it is possible to improve the performance of the algorithm.}
\end{itemize}

\section{Dataset}

The ADE corpus was introduced by Gurulingappa \emph{et al.} 
\shortcite{Gurulingappa2012} in order to provide a benchmark dataset for the 
development of algorithms for the detection of ADRs in case reports. The 
original source of the data was 2972 MEDLINE case reports. The data was 
labelled by three trained annotators and their annotation results were 
consolidated into a final dataset including 6728 ADE relations (in 4272 
sentences), as well as 16688 non-ADR relevant sentences.

The authors calculated Inter-Annotator Agreement (IAA), using F1 scores as a 
criterion, for adverse event entities between 0.77 and 0.80 for partial matches 
and between 0.63 and 0.72 for exact matches. For more detail, the reader can 
refer to the work of Gurulingappa \emph{et al.} \cite{Gurulingappa2012}.

\subsection{Preprocessing}

The dataset is suitable for two types of tasks: (1) categorization of sentences 
as either relevant for ADRs or not; and (2) extraction of drug-adverse event 
relations and drug-dose relations. Because there can be more than one relation 
in the same sentence, the ADR relevant sentences are sometimes duplicated.

The presence of duplicates can lead to situations where the same sentence is 
present in both the training and test datasets, as well as to an overall 
distortion of the distribution of the sentences. In order to prevent this, we 
de-duplicate these sentences, which results in 4272 ADR relevant sentences, as 
stated in the work of Gurulingappa \emph{et al.} \cite{Gurulingappa2012}. 

\section{Methods}

In the following sections, we will describe (1) the word embeddings used in our 
learning algorithms; and (2) the two different CNN architectures evaluated in 
our experiments.

\subsection{Embeddings}

\subsubsection*{GloVe 840B}

As in Huynh's work \cite{Huynh2016}, we use pre-trained word embeddings. Huynh 
focused mainly on the general purpose GloVe Common Crawl 840B, 300 dimensional
word embeddings \cite{Pennington2014}. 

\subsubsection*{Pyysalo's Embeddings}

We also evaluate the use of 200 dimensional word2vec embeddings introduced by 
Pyysalo \emph{et al.} \cite{Pyysalo2013}. These word embeddings were fitted on 
a corpus combining PubMed abstracts, PubMed Central Open Access (PMC OA) full 
text articles as well as Wikipedia articles. We also initialize zero valued 
vectors for the unknown word symbol as well as for the padding symbol. 

\subsubsection*{Preprocessing}
    
As in Huynh's work, no new word vectors are initialized for tokens not present 
in the pre-trained vocabulary, and only the tokens that are in the 20000 most 
frequent words in the dataset are included. The remaining tokens are mapped to 
the unknown word symbol vector. We enable the algorithm to optimize the 
pre-trained weights after initialization. We follow the preprocessing strategy 
used by Huynh \cite{Huynh2016}, which is itself based on that of Kim 
\cite{Kim2014}, and includes expansion of contractions, and additionally, all 
non-alphabetic characters are replaced with spaces prior to tokenization.

\section{Convolutional Neural Network Architectures}

In all architectures described below, the sentences are mapped to a vector 
representation, $\mathbf{v}$. Dropout is applied to $\mathbf{v}$ during training
with a dropout probability of 0.5. As in usual classification tasks, the 
predicted probability of a possitive outcome, that is, of the sentence being 
ADR relevant, is given by 

\begin{equation}
    \hat{y} = \rho\Big(\mathbf{v}^{T}\mathbf{w} + b\Big), 
\end{equation}

where $\mathbf{w}$ is a vector of coefficients, $b$ is the intercept, and 
$\rho$ is the sigmoid function.

The objective function to be optimized is the cross entropy, which can also be 
interpreted as an average negative log-likelihood, and is given by 

\begin{multline}
    L\Big( \Theta \Big) = - \frac{1}{N} \Big[ \sum_{i = 1}^{N} y_i log\Big( 
        \hat{y_i} \Big) + \\ (1 - y_i) log\Big( 1 - \hat{y_i} \Big) \Big].
\end{multline}

\subsubsection*{Huynh's CNN architecture}

\begin{figure} 
    \includegraphics[width=0.42\textwidth]{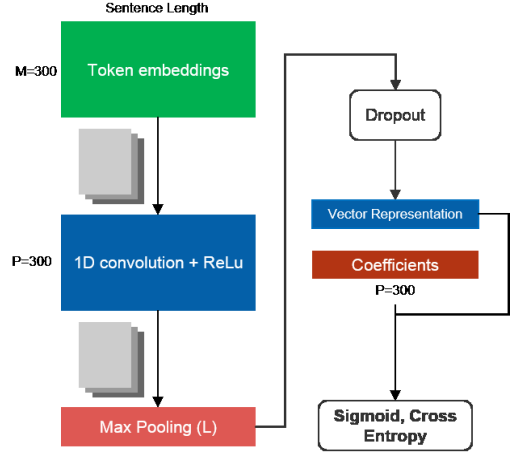}
    \caption{Diagram of the architecture proposed by Huynh \cite{Huynh2016}.}
    \label{fig:huynh_cnn}
\end{figure}

This architecture consists of the use of a 1D-convolution layer with 300 filters
and a 5 token window applied on the word vectors. This is followed by a 
Rectified Linear Unit (ReLu) and a 1D-max pooling over the full axis of 
1D-convolution results. This leads to a 300 dimensional vector representation, 
$\mathbf{v}$, which is used as an input for the classification network 
described above. Figure \ref{fig:huynh_cnn} shows a diagram of the resulting 
architecture. Note that $M$, the number of embedding dimensions, may be equal 
to either 300 or 200, but is shown as 300 for illustration in the figure.

To reduce overfitting, a constraint is added to ensure that the $L_2$ norms of 
each one of the 1D convolution filters are never above a threshold value, $s$, 
after each batch. For more detail, the reader can refer the works of Huynh 
\cite{Huynh2016} and Kim \cite{Kim2014}.

\subsubsection*{Hughes' CNN architecture}

\begin{figure} 
    \includegraphics[width=0.42\textwidth]{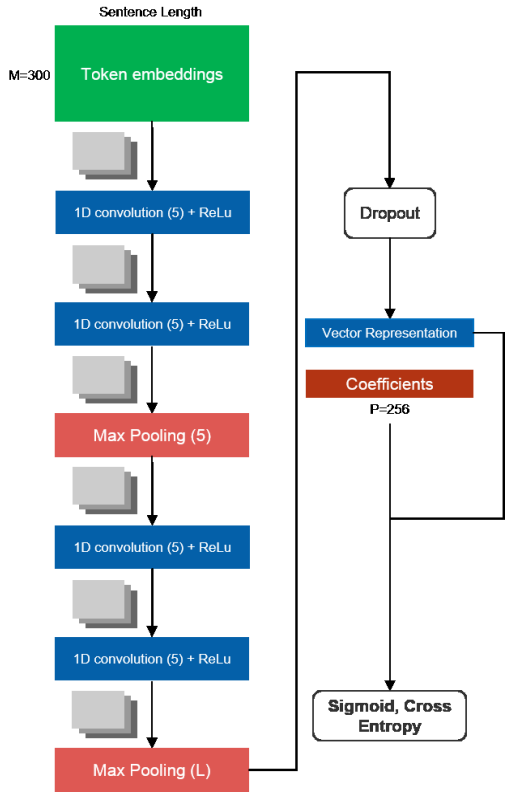}
    \caption{Diagram of an architecture based on the one proposed by Hughes 
    \cite{Hughes2017}.}
    \label{fig:hughes_cnn}
\end{figure}

Based on the approach proposed by Hughes \cite{Hughes2017} we explored a deeper 
architecture, with multiple successive stages of 1D-convolution, non-linear 
transformations, and max pooling.

This architecture starts with two successive stages of 1D-convolutions with 
256 filters and a 5 token window, each followed by a ReLu transformation. After 
this, a 1D-max pooling on the axis of the convolutions with a window of length 
5 is applied. Finally, another two successive stages of 1D-convolutions with 
256 filters and a window of length 5, each followed by a ReLu transformation, 
is applied, followed by a 1D-max pooling over the full axis of the 
1D-convolutions.

Similar to the case of the previous architecture, this leads to a 256 
dimensional vector representation, $\mathbf{v}$, and a constraint is used to 
keep the $L_2$ norms of all 1D-convolution filters under a threshold value $s$.
Figure \ref{fig:hughes_cnn} shows a diagram of the resulting architecture. 
As previously, note that $M$ may be equal to either 300 or 200, but is shown as 
300 for illustration in the figure.

For further detail, the reader can refer to the work of Hughes 
\shortcite{Hughes2017}.

\section{Experimental Setup}

Following the approach used by Huynh \emph{et al.} \shortcite{Huynh2016}, we 
used 10-fold cross validation to evaluate the performance of our classifiers. 
The normalization threshold used to clip the $L_2$ norms of the filters, $s$, 
was set to 9.

The Adam optimizer \cite{Kingma2014} was used to minimize the loss, $L\Big( 
\Theta \Big)$, with 8 epochs and a batch size of 50. To avoid overfitting, 
early stopping is used based on a development set consisting of 10\% of the 
training data of each fold. For the decision of the classifier, instead of a 
$\hat{y}$ threshold of 0.5, we determine the optimum threshold by evaluating 
all possible thresholds present in the development set of each fold and keeping 
the threshold that results in the best F1 score.

After every 10 batches, the optimal threshold is determined from the development
set and the associated best F1 score is obtained. Optimization is stopped if 
the F1 score on the development set fails to improve after 6 steps. The set of 
CNN parameters associated with the best F1 score observed throughout the 
training process is then kept and used to evaluate the network's performance on 
the test set of each fold. 

We use the architecture originally proposed by Huynh \cite{Huynh2016} without 
de-duplication as the baseline results to understand the impact of the 
de-duplication, choice of embeddings, and CNN architecture.

All CNN implementations were done using Python 3.4.5 \cite{Python} and 
Tensorflow 1.2.0 \cite{Tensorflow2015}.

\section{Results}

\subsection{Impact of De-duplication on Classification Performance Estimates}

\begin{table}[h]
\begin{center}
\begin{tabular}{|l|r|r|}
\hline \bf De-duplication & \bf No & \bf Yes \\
\hline
Accuracy & 0.919 & 0.914 \\
Precision & 0.858 & 0.784 \\
Recall & 0.860 & 0.798 \\
F1-score & 0.859 & 0.790 \\
Specificity & 0.942 & 0.943 \\
AUROC & 0.966 & 0.954 \\
\hline
\end{tabular}
\end{center}
\label{tab:deduplication}
\caption{Performance metrics of Huynh's architecture using GloVe 840B 
embeddings with and without de-duplication of the ADR relevant sentences.}
\end{table}

Table \ref{tab:deduplication} shows a comparison of the performance metrics of 
our implementation of Huynh's architecture and GloVe 849B word embeddings with 
and without de-duplication of the sentences labelled as ADR relevant. After 
de-duplication, most of the performance metrics were lower, since the presence 
of duplicates in the positive samples resulted in overly optimistic results.

The biggest impact was observed on precision, recall and F1 scores. Overall 
accuracies and area under the ROC curve (AUROC) didn't seem to be greatly 
affected. Note that the specificity, which is the true negative rate, was 
higher after de-duplication.

We initially obtained somewhat lower performances for the baseline model 
without de-duplication compared to the one reported by Huynh \emph{et al.} 
\shortcite{Huynh2016} even though we accurately followed the described 
architecture. After investigating the differences in the code, we noticed that 
during pre-processing, characters that are not alphabetic are replaced with 
spaces prior to tokenization. After incorporating this step into our code, the 
results matched the previously reported ones much better.

\subsection{Impact of Biomedical Word Embeddings}

\begin{table}[h]
\begin{center}
\begin{tabular}{|l|r|r|}
\hline \bf Word Embeddings & \bf Glove 840B & \bf Pyysalo \\
\hline
Accuracy & 0.914 & \bf 0.918 \\
Precision & 0.784 & \bf 0.800 \\
Recall & \bf 0.798 & 0.797 \\
F1-score & 0.790 & \bf 0.798 \\
Specificity & 0.943 & \bf 0.949  \\
AUROC & 0.954 & \bf 0.958 \\
\hline
\end{tabular}
\end{center}
\label{tab:embeddings}
\caption{Performance metrics of Huynh's architecture with de-duplication with 
GloVe 840B embeddings and Pyysalo's embeddings.}
\end{table}

Table \ref{tab:embeddings} shows a comparison of the performance metrics with 
de-duplication of ADR relevant sentences using the GloVe 840B word embeddings, 
and the word embeddings fit for biomedical data purposes proposed by Pyysalo 
\emph{et al.} \cite{Pyysalo2013}. 

In most cases, the use of biomedical word embeddings was favorable or 
non-detrimental to the performance metrics. The largest improvement was seen on 
the increase of average precision from 0.780 with GloVe 840B to 0.800 with the 
biomedical embeddings. 

This also led to an increased average F1 score from 0.790 to 0.798. The average 
AUROC also increased from 0.954 to 0.958. Specificity increased from 0.943 to 
0.949, and recall was the only metric that was slightly reduced 
from 0.798 to 0.797.

\subsection{Comparison With Hughes' CNN Architecture}

\begin{table}[h]
\begin{center}
\begin{tabular}{|l|r|r|}
\hline \bf Architecture & \bf Huynh & \bf Hughes \\
\hline
Accuracy & \bf 0.918 & 0.905 \\
Precision & \bf 0.800 & 0.765 \\
Recall & \bf 0.797 & 0.771 \\
F1-score & \bf 0.798 & 0.767 \\
Specificity & \bf 0.949 & 0.939 \\
AUROC & \bf 0.958 & 0.940 \\
\hline
\end{tabular}
\end{center}
\label{tab:architectures}
\caption{Performance metrics of Huynh's and Hughes' architectures with 
de-duplication and Pyysalo's embeddings.}
\end{table}

Table \ref{tab:architectures} shows a comparison between the performances of 
our implementations of Huynh's and Hughes' architectures. In both cases, 
de-duplication of ADR relevant sentences, and biomedical embeddings were used. 
The former ourperformed the latter in every performance metric. The biggest 
improvement was in metrics associated to the positive class, such as precision, 
recall, and F1 score.

\section{Discussion}

The purpose of this work was to evaluate the use of convolutional neural 
networks (CNNs) architectures and biomedical word embeddings for the automatic 
categorization of sentences relevant to adverse drug reactions (ADRs) in 
case reports present in the biomedical literature. For this purpose, we used 
the ADE corpus, which consists of sentences coming from 2972 MEDLINE case 
reports labelled by trained annotators. This includes 4272 ADR relevant 
sentences, as well as 16688 non-ADR relevant sentences.

We showed that, because of duplications present in the ADE corpus, the use of 
this dataset for sentence classification without performing a de-duplication 
can lead to overoptimistic performance estimates. In addition, we showed that, 
by using biomedical word embeddings, as opposed to general purpose word 
embeddings, it's possible to improve upon the performance of the algorithm. 
Finally, we compared the performance of our implementations of two CNN 
architectures, with the architecture proposed by Huynh outperforming the 
architecture proposed by Hughes in this task and dataset in every metric.

One important measure of the potential noise in the inputs of human annotators 
is the Inter Annotator Agreement (IAA) \cite{Gurulingappa2012}, which in this 
dataset was measured by its original authors by calculating inter annotator 
F1 scores. Although this measure was calculated on the entity (partial and 
exact) matching level, and although there has been a harmonization process, 
it is informative of the potential noise in the inputs used to build the 
dataset. The fact that the IAAs for partial matches of adverse events ranged 
between 0.77 and 0.80 indicates that aiming for near perfect predictions may 
be unrealistic, since there is a considerable degree of disagreement between 
human annotators.

\section{Conclusions and Future Work}

Our results highlight the importance of sentence de-duplication, pre-processing,
choice of word embeddings, and neural network architectures when applying 
convolutional neural networks (CNNs) for the detection of adverse drug reaction 
(ADR) relevant sentences in the biomedical literature using the ADE dataset. 
We believe that these are only a few of the factors that can greatly influence 
the performance of the algorithms performing these tasks.

Future work could include the use of either exhaustive, grid-based or 
reinforcement-learning based search for more optimal CNN architectures, as well 
as the evaluation of architectures other than CNNs. In addition, another very 
interesting area explored in previous works \cite{Huynh2016} was the aspect of 
visualization using CNNs with Attention (CNNAs). However, this algorithm seemed 
to underperform compared to the normal CNN. Building upon this approach to 
improve its performance while retaining its attractive visualization properties 
would be an important step towards the development of systems that assist 
human readers.

\section{Acknowledgements}

The author would like to thank Abhimanyu Verma as well as the Technology 
Architecture \& Digital department at Novartis Pharma A.G. for their support in 
this research.

\bibliography{acl2017}
\bibliographystyle{acl_natbib}

\appendix

\end{document}